\documentclass[10pt,twocolumn,letterpaper]{article}

\usepackage[pagenumbers]{wacv} % To force page numbers, e.g. for an arXiv version

\usepackage{graphicx}
\usepackage{amsmath}
\usepackage{amssymb}
\usepackage{booktabs}
\usepackage{multirow}
\usepackage{multicol}
\usepackage[table]{xcolor} % This includes the xcolor package with table option
\usepackage{colortbl} % This includes the colortbl package
\usepackage{xcolor}
\usepackage{color}
\usepackage{times}
\usepackage{verbatim}
\usepackage{array}
\usepackage{float}
\usepackage{multirow}
\usepackage{multicol}
\usepackage{siunitx}
\usepackage{booktabs}
\usepackage{colortbl}
\usepackage{enumitem}
\usepackage{listings}
\usepackage{tablefootnote}
\usepackage{latexsym}
\usepackage{url}
\definecolor{newcolor}{rgb}{.8,.349,.1}
\usepackage{times}
\usepackage{epsfig}
\usepackage{graphicx}
\usepackage{colortbl}
\usepackage{float}
\usepackage{algpseudocode}
\usepackage[ruled]{algorithm2e}
\usepackage{caption}
\usepackage{etoolbox}
\usepackage[normalem]{ulem}
\usepackage{csquotes}
\usepackage{amsmath}

\definecolor{mygray}{gray}{0.9}

\usepackage{tcolorbox}
\newtcolorbox[list inside=prompt,auto counter,number within=section]{prompt}[1][]{
    colbacktitle=black!60,
    coltitle=white,
    fontupper=\footnotesize,
    boxsep=5pt,
    left=0pt,
    right=0pt,
    top=0pt,
    bottom=0pt,
    boxrule=1pt,
    title={#1},
    #1, % add more args
}

\usepackage[pagebackref,breaklinks,colorlinks]{hyperref}

\usepackage[capitalize]{cleveref}
\crefname{section}{Sec.}{Secs.}
\Crefname{section}{Section}{Sections}
\Crefname{table}{Table}{Tables}
\crefname{table}{Tab.}{Tabs.}

\def\wacvPaperID{1405} % *** Enter the WACV Paper ID here
\def\confName{WACV}
\def\confYear{2025}

\begin{document}

%%%%%%%%% TITLE - PLEASE UPDATE
\title{MIP-GAF: A MLLM-annotated Benchmark for Most Important Person Localization and Group Context Understanding}

\author{Surbhi Madan\\
IIT Ropar\\
% Institution1 address\\
% {\tt\small surbhi.19csz0011@iitrpr.ac.in}
% For a paper whose authors are all at the same institution,
% omit the following lines up until the closing ``}''.
% Additional authors and addresses can be added with ``\and'',
% just like the second author.
% To save space, use either the email address or home page, not both
\and
Shreya Ghosh\\
Curtin University, Australia\\
% {\tt\small secondauthor@i2.org}
\and
Lownish Rai Sookha\\
IIT Ropar\\
% {\tt\small secondauthor@i2.org}
\and
M.A. Ganaie\\
IIT Ropar\\
% {\tt\small secondauthor@i2.org}
\and
Ramanathan Subramanian \\
University of Canberra,Australia\\
\and
Abhinav Dhall\\
Flinders University, Australia\\
\and
Tom Gedeon\\
Curtin University, Australia\\
% {\tt\small secondauthor@i2.org}
}

\twocolumn[{%
\renewcommand\twocolumn[1][]{#1}%
\maketitle
\begin{center}
    \centering
    \captionsetup{type=figure}
    \includegraphics[width=0.85\linewidth,height=105mm]{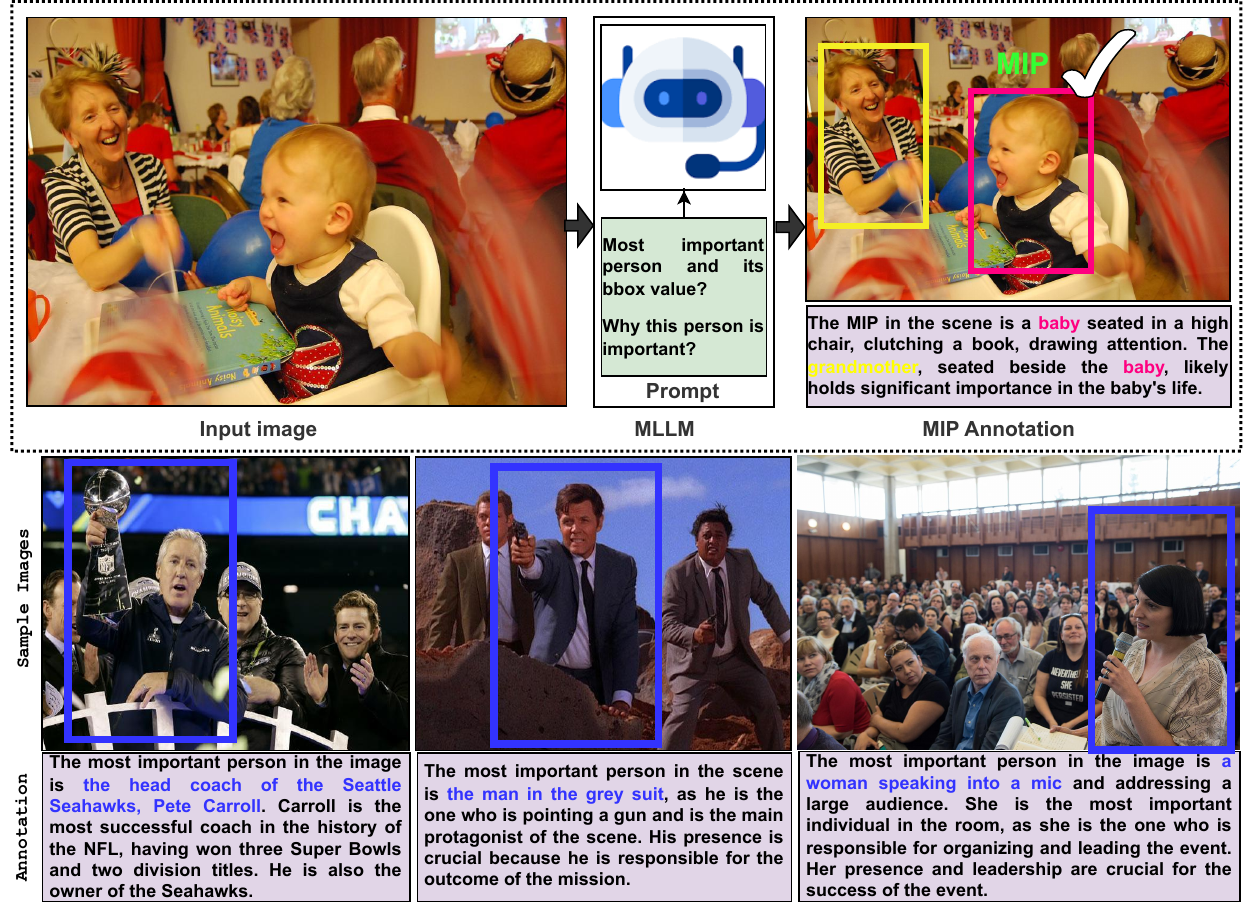}
    % \captionof{figure}{Test caption}
     \caption{The MIP-GAF is an image-based dataset which contains the location of the most important person along with its context-based understanding and explanation. \textbf{Top:} This part shows an overview of MIP-GAF dataset annotation with a multimodal large language model. \textbf{Bottom:} Sample annotated images. \textit{Left.} This image is the celebration of a match in which the person holding the trophy is the most important person. \textit{Middle.} In an action-based movie scene, the person holding the gun is the MIP here. \textit{Right.} In this image, given a wide audience, the
women speaking is the most important person.}
\label{fig: teaser}
\end{center}%

}]

%%%%%%%%% ABSTRACT
\begin{abstract}
 Estimating the Most Important Person (MIP) in any social event setup is a challenging problem mainly due to contextual complexity and scarcity of labeled data. Moreover, the causality aspects of MIP estimation are quite subjective and diverse. To this end, we aim to address the problem by annotating a large-scale `in-the-wild' dataset for identifying human perceptions about the `Most Important Person (MIP)' in an image. The paper provides a thorough description of our proposed Multimodal Large Language Model (MLLM) based data annotation strategy,  and a thorough data quality analysis. Further, we perform a comprehensive benchmarking of the proposed dataset utilizing state-of-the-art MIP localization methods, indicating a significant drop in performance compared to existing datasets. The performance drop shows that the existing MIP localization algorithms must be more robust with respect to `in-the-wild' situations. We believe the proposed dataset will play a vital role in building the next-generation social situation understanding methods. The code and data is available at \url{https://github.com/surbhimadan92/MIP-GAF}.
\end{abstract}
\vspace{-0.2 in}
%%%%%%%%% BODY TEXT
\section{Introduction}
% ******Original Table*******************
\begin{table*}[t]
\caption{Details of available MIP detection datasets are in chronological order. * denotes datasets having unlabelled samples.}
\label{tab:datasets}
\centering
\scalebox{0.7}{
\begin{tabular}{l||c|c|c|c|c|c|c|l}
\toprule[0.4mm]
\rowcolor{mygray}\textbf{Dataset} & \textbf{Year} & \textbf{\#Total} & \multicolumn{4}{c|}{\textbf{\#Stat}} & \textbf{Anno.} & \textbf{Scene}\\ \cline{4-7}
\rowcolor{mygray}\textbf{} & \textbf{}& \textbf{} & \multicolumn{2}{c|}{\textbf{Train}} & \textbf{Val} & \textbf{Test} & \textbf{}  & \textbf{} \\ \cline{4-7}
\rowcolor{mygray}\textbf{} & \textbf{} & \textbf{} & \textbf{Labelled} & \textbf{Unlabelled} & \textbf{} & \textbf{} & \textbf{}  & \textbf{} \\
\hline\hline
GAF-Personage~\cite{ghosh2018role} & 2018 & 1,000 & NA & NA & NA &NA & Manual & NA\\
MS Dataset~\cite{li2018personrank} & 2018 & 2,310 & 924 & NA & 232 & 1,154 &  Manual & Speech, Demonstration, Interview, Sports, Military, Meeting \\
NCAA Basketball~\cite{li2018personrank, ramanathan2016detecting} & 2018 & 9,736 & NA & NA & NA & NA & Manual & Basketball\\
EMS Dataset~\cite{li2019learning} & 2020 &  10,687* & 690&8,377 &230 &1,390 & Manual & Speech, Demonstration, Interview, Sports, Military, Meeting\\
ENCAA Basketball~\cite{li2019learning} & 2020 & 19,062* & 2,825 & 19,065 & 941 & 5,970 & Manual &  Basketball\\
Unconstrained-7k~\cite{wang2021very} & 2021 & 7,250 & 3,625& NA & NA &3,625 & Manual & Speech, Demonstration, Interview, Sports, Military, Meeting\\
CUC Dataset~\cite{wang2022towards} & 2022 & 9,390 & 4,694 & NA & NA & 4,696 & Manual  & Speech, Demonstration, Interview, Sports, Military, Meeting \\ \hline
 \textbf{MIP-GAF (Ours)} & 2024 & \textbf{16,550} &9,615 & NA &4,113 & 2,822 & \shortstack{Semi- \\ automatic}  & 
 \begin{tabular}{l}
\textbf{Casual Gathering:} Family get-together, Friends \\ conversation, Festivals \\
\textbf{Celebration:} Birthday, Crowd cheering, Match winning\\
\textbf{Fighting:} Street fight, Boxing, Crowd fighting, Fighting\\
\textbf{Group Activities:} Community service, People on street,\\  Religious gathering, Classes, Group dance, March-past\\
\textbf{Funeral:} Condolence meeting\\
\textbf{Meeting:} Event announcement, Group discussion,\\ Interview, Conversation, Discussion, Press conference\\ 
\textbf{Protest:} Stone pelting, Violent Protest, Protest
Quarrelling \\ 
\textbf{Show:} Concert, Live shows, TV shows, Talk shows\\
\textbf{Sports:} People watching match, Wrestling
 \end{tabular}  \\ 
\bottomrule[0.4mm]
\end{tabular}}
\vspace{-6mm}
\end{table*}

 The localization of a \textit{Most Important Person (MIP)} in a multi-person social scene provides important cues related to a range of real-world applications such as image captioning~\cite{tang2021clip4caption, ma2023towards}, social relation analysis~\cite{sun2017domain}, group activity recognition~\cite{ghosh2020automatic, gavrilyuk2020actor}, group emotion analysis~\cite{xie2023most,ghosh2020automatic,dhall2017individual} and dominant person in group~\cite{sharma2023graphitti, zhao2021space}. MIP estimation in unconstrained environments (as shown in Figure~\ref{fig: teaser}) is quite challenging due to higher-order relationships among scene objects and human(s), situational impact, camera position, occlusion, blur, and presence of multiple people. Among the challenges mentioned above, encoding a higher-order relationship between objects and the scene is challenging, requiring precise detection of objects and humans and encoding their interactions. Similarly, the camera position is essential in constructing the scene-level subjective perception. `Importance' among subjects and objects in a still frame is ambiguous. Usually, a still image has many perspectives, such as the photographer's point of view, social norms, and viewers' (third person) perspective. While capturing any scene, a photographer aims to capture some `important' aspect in that still frame. Thus, the main aim of the photographer remains either unknown or non-visible to the viewer. The camera angle also plays a crucial role in the perception of importance. The human visual system first focuses on the most significant foreground object(s)/subject(s) in an image instead of the background. Similarly, the relative position of people in an image plays an important role in the perception of `important people'. For example, in social events, the important person is likely to be centered in the frame. However, these aspects can greatly vary according to social norms, context, and informal settings. In cases where a socially prominent personality is present in the scene, they are presumed to be the most important. Predicting the important person in images is therefore a challenging task.

The problem of detecting the MIP is two-fold in terms of contribution, \textit{i.e.}, data availability and localization. The development of data-centric deep learning algorithms~\cite{singh2023systematic} depends on the quality of the annotated data. The context understanding aspect in localization of MIP has been overlooked in the literature~\cite{ghosh2018role,li2018personrank,hong2020learning,li2019learning}. The MIP localization is a complex problem as more reasoning-based perception is involved instead of a simple object detection/localization perspective~\cite{zhai2024background}. Localizing MIP also involves ranking aspects amongst the people present in the image, which poses an extra challenge to the problem regarding the number of people, their visibility, resolution, and camera perspective. Also, in certain images, there are `no MIPs' or `multiple MIPs' based on the third person's perspective. These images introduce noise to the learning protocol.

% To this end, we address these gaps by releasing a new large-scale, `in-the-wild' dataset for detecting the most important person in an image, and designed explicitly with the reasoning of the `ground truth' context.

To address these gaps, we are releasing a new large-scale ``in-the-wild" dataset for detecting the most important person in an image, explicitly designed with ``ground truth" context reasoning. With the surge in large language models~\cite{kasneci2023chatgpt,chang2023survey}, the contextual reasoning of an image has become quite popular. Thus, we initialize the data annotation pipeline with multimodal large language models or MLLM-based models before manual validation. In summary, our contributions are:
\begin{itemize}[topsep=1pt,itemsep=0pt,partopsep=1ex,parsep=1ex,leftmargin=*]
    \item We propose MIP-GAF (Group AFfect), a large-scale MLLM-driven MIP localization benchmark that covers the reasoning aspects of people interacting in an image. Please note that we annotate images from a third-person perspective following the literature.
    \item We incorporate a novel semi-automatic MLLM-based data annotation strategy, which covers the context-based reasoning aspect of localizing the most important person.
    \item We perform a comprehensive analysis and benchmark the proposed MIP-GAF dataset using state-of-the-art MIP detection algorithms, including the proposed MIP-CLIP benchmark. We evaluate the datasets across four learning paradigms: zero-shot, fully supervised, semi-supervised, and self-supervised. The significant performance drop ($\sim \textbf{19.21}$ mAP for MS dataset and $\sim \textbf{24.21}$ mAP for NCAA dataset with the supervised POINT framework) indicates that our dataset will be a valuable asset to pursue further research in this domain. Additionally, current methods perform better when trained on MIP-GAF.

    % \item We perform a comprehensive analysis and benchmark the proposed dataset using state-of-the-art MIP detection algorithms alongside the proposed MIP-CLIP benchmark. We benchmark the datasets for four different learning paradigms, \textit{i.e.}, zero-shot, fully supervised, semi-supervised, and self-supervised. The significant drop in performance (i.e., $\sim \textbf{19.21}$ mAP drop w.r.t MS dataset and $\sim \textbf{24.21}$ mAP drop w.r.t NCAA dataset with supervised POINT framework) directly indicate that our dataset will be a valuable asset to pursue further research in this domain. We also observe that current methods also perform better when trained on our MIP-GAF dataset.
\end{itemize}

% \begin{figure*}[t]
%     \centering
%     \includegraphics[width = \linewidth]{images/kosmos2-sample-1.pdf}
%     \caption{Sample output from the Kosmos-2 model. \textit{Left.} In this image, the girl holding the cake is the most important person given the friendly gathering context. \textit{Middle.} In a action based movie scene, the person holding the gun is the MIP here. \textit{Right.} In this image, it is the celebration of a match where the person holding the trophy is the most important person. }
%     \label{fig:kosmos2}
%     \vspace{-3mm}
% \end{figure*}
% \vspace{-4mm}

%------------------------------------------------------------------------
\section{Related Work}
This section reviews research on (a) Localizing the MIP, (b) MIP datasets and (c)  Group context understanding to position our work with respect to the literature.

\noindent \textbf{Localizing the Most Important Person.}
Localizing the important person/object in egocentric videos is a well-explored problem in the literature~\cite{huang2023egocentric, yagi2018future,lee2012discovering}. Instead, we explore the problem from a third-person point of view, which is highly relevant to the studies of important person detection given any image~\cite{li2018personrank,ghosh2018role}. The prior work mainly focuses on developing either supervised algorithm~\cite{li2018personrank,ghosh2018role} or semi/unsupervised method~\cite{hong2020learning} to predict the most important person. In particular, Ghosh et al.~\cite{ghosh2018role} propose a coarse-to-fine multiple instance learning strategies for important person detection; Li et al.~\cite{li2018personrank} build a hybrid graph modeling the interaction among persons in the image and develop a graph model called PersonRank to rank the individuals in terms of importance scores from the hybrid graph. Further, Li et al.~\cite{li2019learning} propose an end-to-end network, POINT, that can automatically learn the relations among individuals. Among learning with less supervision paradigms, Hong et al.~\cite{hong2020learning} mainly focus on designing a semi-supervised method to detect the most important person while taking advantage of the unlabelled data. Similarly, GraphITTI~\cite{sharma2023graphitti} proposes a homogeneous attributed graph framework to predict the most dominant person in a group interaction setting. To the best of our knowledge, we benchmark the MIP-GAF dataset from four different settings: zero-shot, supervised, semi-supervised, and self-supervised with diverse and challenging contexts.

\noindent \textbf{Most Important Person Localization Datasets.}
A comprehensive comparison of available datasets in the MIP domain is presented in Table~\ref{tab:datasets}. The prior works~\cite{ghosh2018role, li2018personrank} have collected several small-scale datasets i.e., \textit{GAF Personage}~\cite{ghosh2018role}, \textit{Multi-scene Important People Image Dataset (MS dataset)}~\cite{li2018personrank} and  \textit{NCAA Basketball Image Dataset}~\cite{li2018personrank} to facilitate research in the domain of localizing MIP. The small-scale data curation directly indicates the difficulty level in the MIP annotation process. The MS dataset has mined the images from the web having different `event + person' tags such as \textit{lecture/speech}, \textit{demonstration}, \textit{interview}, \textit{sports}, \textit{military} and \textit{meeting}. At the same time, the NCAA dataset contains different event images of \textit{basketball} game. Later, Hong et al.~\cite{hong2020learning} have released extended versions of NCAA and MS datasets in a semi-supervised way, i.e., \textit{ENCAA}, and \textit{EMS}. The context information of the NCAA~\cite{li2019learning} and ENCAA~\cite{li2019learning} datasets are simple, including only \textit{basketball} sports scenes. The reasoning for identifying the most important person (MIP) is consistent, focusing on \textbf{key players} interacting with the ball, either by shooting or holding it. MS~\cite{li2018personrank} and EMS~\cite{li2018personrank} datasets are relatively rich in types of scenes under constrained conditions. Mostly, the MIPs are either in frontal view concerning the camera or fall under salient regions of the image. These datasets are biased toward uniform settings, and the algorithms' effectiveness may be impacted in unconstrained situations. To address this limitation, the Unconstrained-7k dataset~\cite{wang2021very}, which contains 7,250 annotated images from various unconstrained scenes, is proposed. However, it only includes one VIP per image. To overcome this, the CUC dataset reorganizes the MS~\cite{li2018personrank} and Unconstrained-7k~\cite{wang2021very} datasets, incorporating scenarios with both no VIPs and multiple VIPs. In this work, we cover various contexts and the pre-existing factors for determining the most important person in `in-the-wild' situations.
% To overcome such scenarios, the Unconstrained-7k dataset~\cite{wang2021very} containing 7,250 annotated images covering various scenes under unconstrained conditions is proposed. However, the number of VIPs in the Unconstrained-7k dataset is limited to one. To overcome this limitation, the \textit{CUC dataset} reorganized the MS~\cite{li2018personrank} and Unconstrained-7k~\cite{wang2021very}. CUC dataset~\cite{wang2022towards} also includes situations containing \textit{no VIPs} and \textit{multiple VIPs}. Here, we cover various contexts and the pre-existing factors for determining the most important person in `in-the-wild' situations.

\begin{figure*}[t]
    \centering
    \includegraphics[width=39mm,height=40mm]{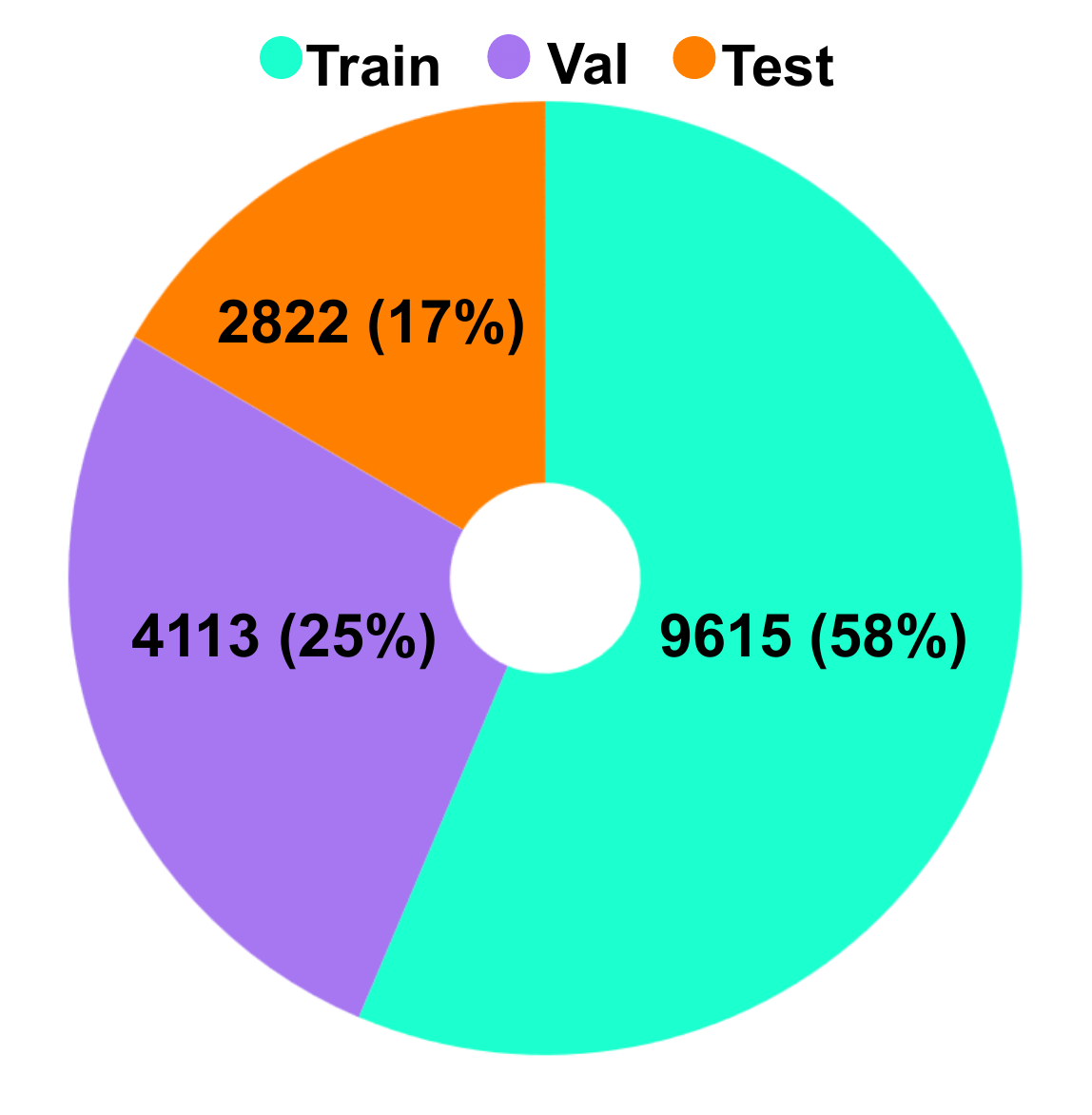}
    \includegraphics[width=49mm,height=40mm]{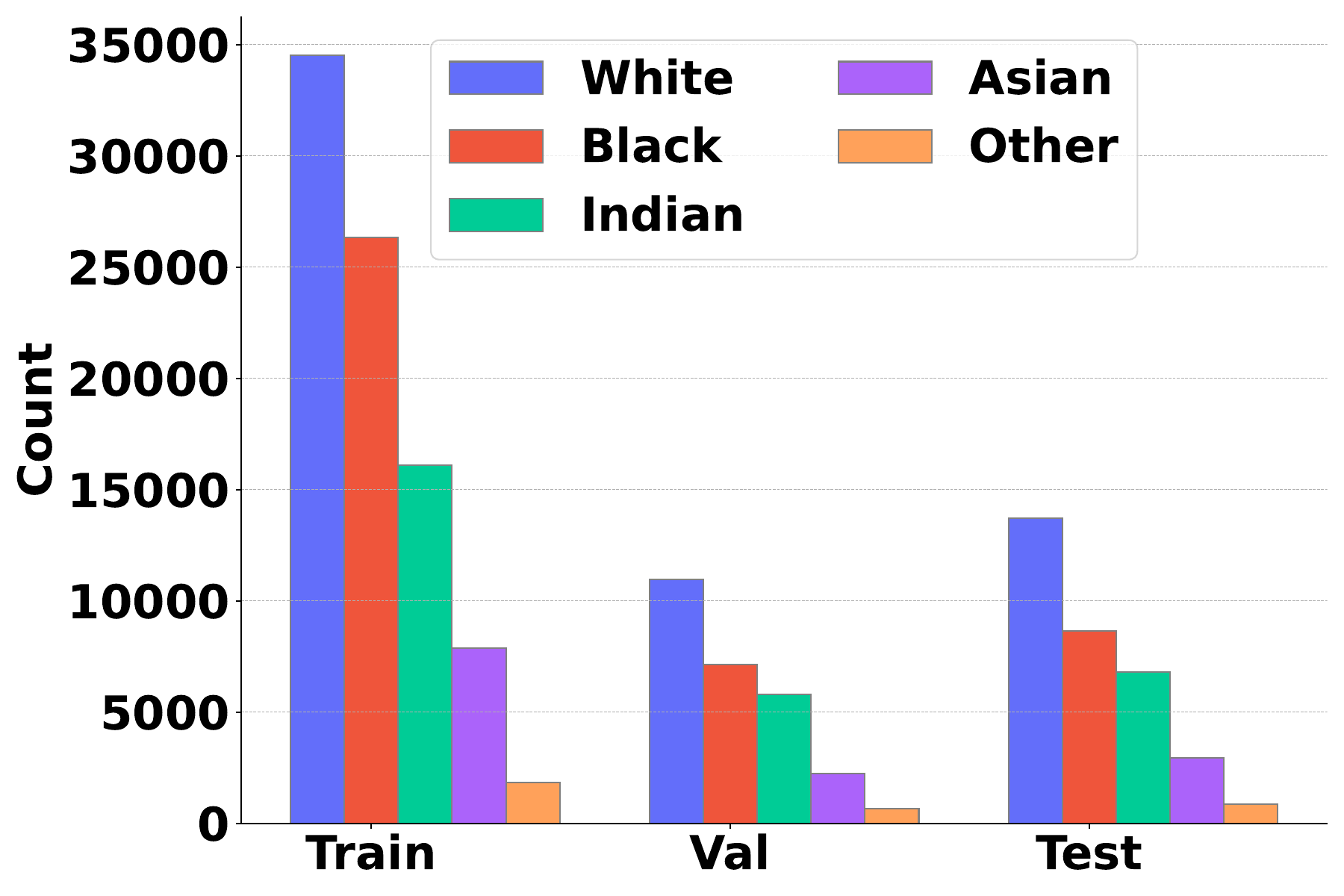}
    \includegraphics[width=39mm,height=40mm]{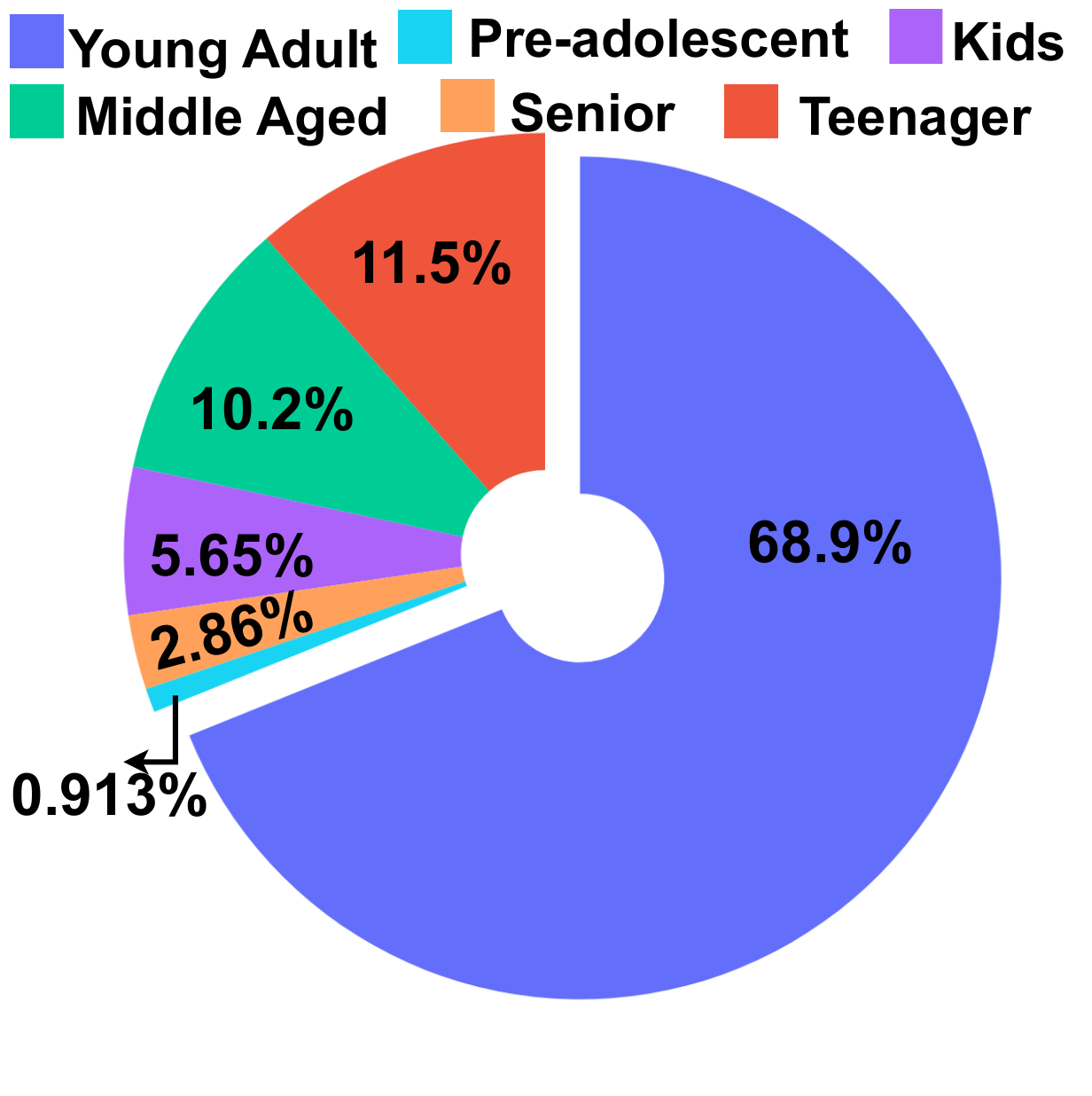}
    \includegraphics[width=45mm,height=40mm]{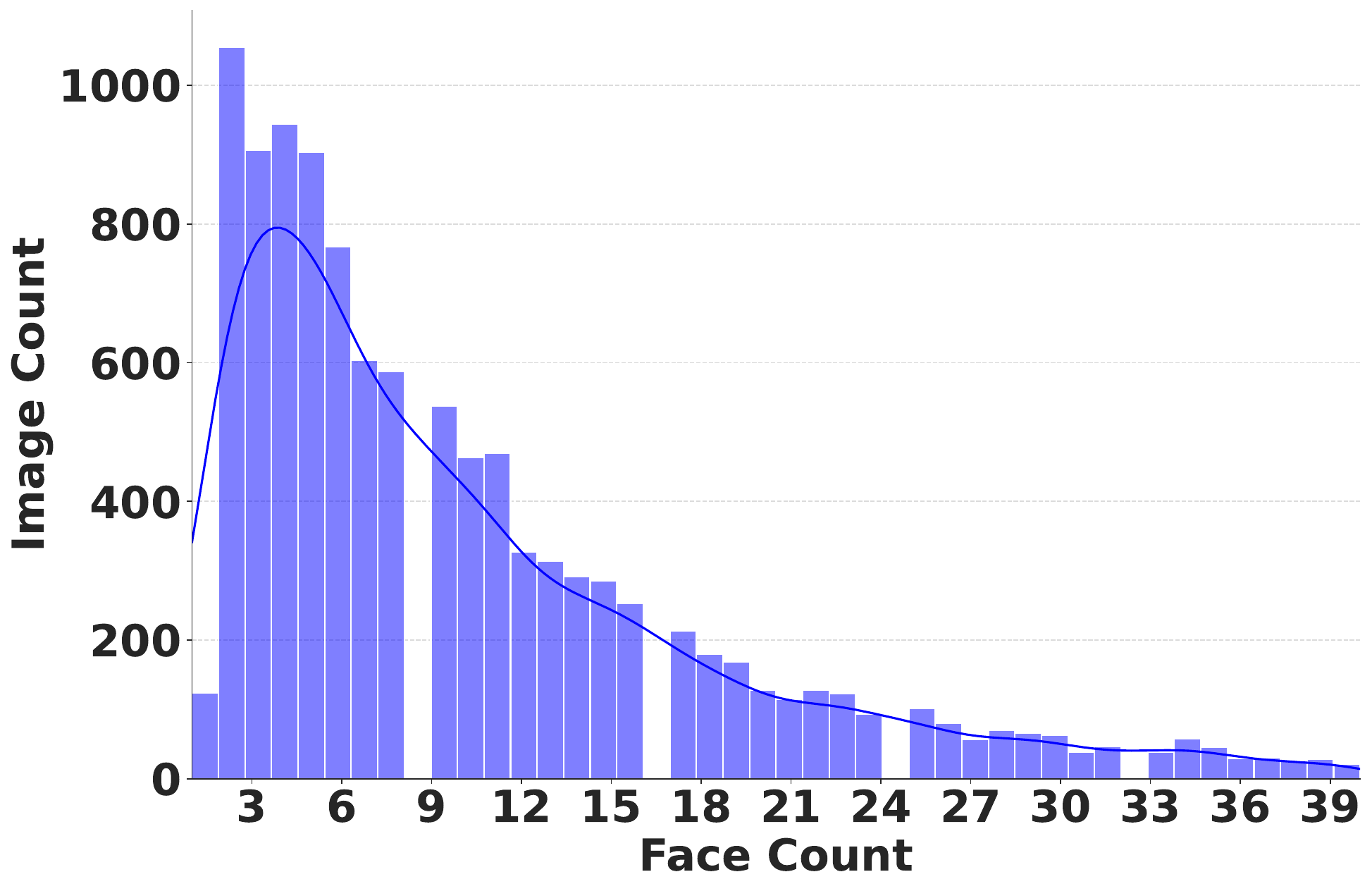}
    \caption{Overview of data statistics. \textit{Left.} Overview of train, validation and test set splits. \textit{Second Left.} Ethnicity distribution over the three splits. \textit{Second Right.} Age distribution of the detected persons. %using~\href{https://github.com/vladmandic/face-api?tab=readme-ov-file}{Face-api}. 
    \textit{Right.} Per image detected face distribution.}
    \label{fig:data_stat_fig}
    \vspace{-6mm}
\end{figure*}

\noindent \textbf{Group Context Understanding.} Prior studies include a range of non-verbal cues and context information associated with MIP analysis. Yamaguchi et al.~\cite{berg2012understanding} define `importance' via several human perceived factors such as compositions (i.e., size and location of objects), semantics (i.e., object type, scene type along with its description strength) and context of the given image. Modeling of the social interaction between subjects is highly related to MIP. \textit{PersonRank}~\cite{li2018personrank} algorithm utilizes pairwise interaction and hyper interaction features to infer the MIP in an image. In general, in group interaction scenes, non-verbal cues such as eye gaze~\cite{shim2021impact}, head direction~\cite{madan2021head} and gestures~\cite{sharma2023graphitti, madan2023magic} play an important role. Additionally, event information~\cite{lee2019context, wang2022congnn} is analyzed, such as birthday party elements like a cake and relevant objects, or a sports celebration featuring a person holding a trophy. Some examples are shown in Figure~\ref{fig: teaser}.

% Along with that, the context or event information~\cite{lee2019context, wang2022congnn} by analyzing the event-related context such as in a birthday party cake and other relevant objects and a sports celebration person holding the trophy. Some of the examples are shown in Figure ~\ref{fig: teaser}.  

% \subsection{
\noindent \textbf{Analysis of Related Work.}
Upon analyzing related work, our work has the following differences regarding the existing literature.

\begin{itemize}[topsep=1pt,itemsep=0pt,partopsep=1ex,parsep=1ex,leftmargin=*]
    \item Prior datasets mainly deals with small-scale datasets i.e. \textit{GAF Personage}~\cite{ghosh2018role}, \textit{Multi-scene Important People Image Dataset (MS dataset)}~\cite{li2018personrank} and \textit{NCAA Basketball Image dataset} \cite{li2018personrank}. Due to annotation complexity and the subjective nature of annotation, MS and NCAA datasets rely on a very uniform distribution of images. Here, the \textit{MIP}s are either in frontal view concerning the camera or fall under salient regions of the image. In contrast, our MIP-GAF dataset covers diverse `in-the-wild' situations with reasoning aspects.
    % \item We have enough labeled samples for each split, which is not available in other datasets.
    \item We have explanations for every labeled \textit{MIP} showcasing the reason behind their importance. These explanations are aligned with the image context, which plays a vital role in determining the \textit{MIP} from an image. We inject this knowledge into CLIP's text encoder for vision language pre-training and also in proposed MIP-CLIP baseline.
    % \item We also have a corresponding emotion label that can be utilized for emotion recognition.
\end{itemize}
% \vspace{-0.4 in}

%-------------------------------------------------------------------------

\section{MIP-GAF Dataset}
MIP-GAF is a large-scale MIP detection dataset, including 16,550 images containing more than 1,47,044 unique detected subjects captured in diverse background environments. This positions the proposed dataset as the most comprehensive benchmark, as illustrated in Figure~\ref{fig: teaser} and Table~\ref{tab:datasets}. 

% \subsection{Data Annotation Pipeline}

\begin{figure*}[t]
    \centering
    \includegraphics[width=0.85\textwidth]{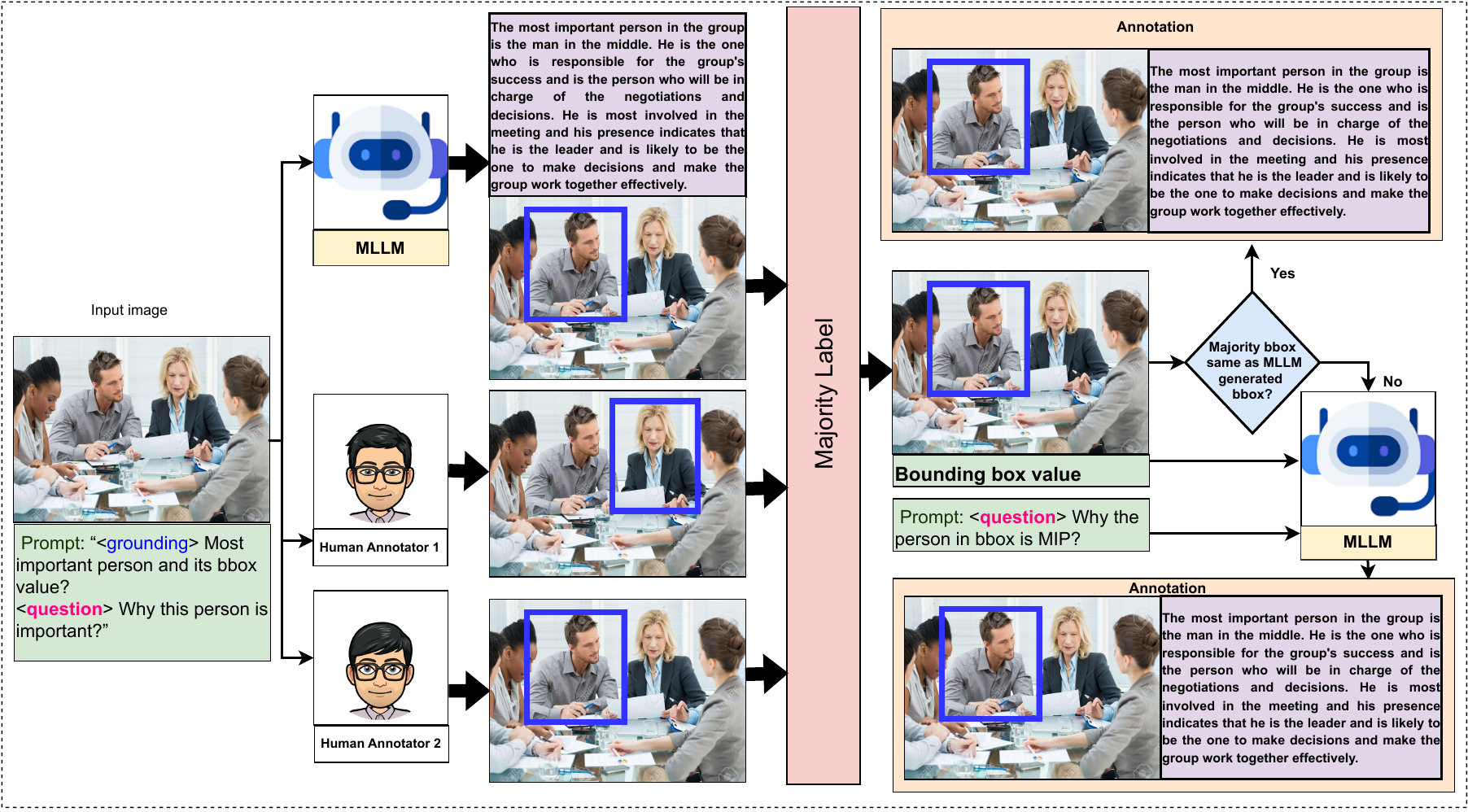}
    \vspace{-3mm}
    \caption{\textit{Data Annotation Pipeline.} Overview of our data labelling paradigm. We bring the concept of `human-in-the-loop' annotation. We initialize the annotation process with MLLM-based annotation followed by a label-refining strategy with human annotators.}
    \label{fig:data_label}
    \vspace{-7mm}
\end{figure*}

\noindent \textbf{Data Annotation Pipeline}
A brief overview of the data annotation pipeline is shown in Figure~\ref{fig:data_label}. In the context of a multi-person image, we utilize the following prompt engineering: 
% ~\\
\begin{prompt}[title={Prompt \thetcbcounter: Kosmos-2 Interaction}]
\textbf{System:} Initialize Kosmos-2.

\textbf{Human:} \{$<$Grounding$>$ Who is the most important person in the given image?\}

\textbf{AI:} \{EXAMPLE OUTPUT is the bounding box of the MIP (See Figure~\ref{fig: teaser})\}

\textbf{Human:} \{and why this person is MIP?\}

\textbf{AI:} \{EXAMPLE OUTPUT is the explanation in text regarding the reason behind the most important person.\} 

The person is important because $<$REASONING$>$.

[('The most important person', (x, y), [(X1, X2, Y1, Y2)])]
\end{prompt}

The prompt engineering mentioned above utilizes the Kosmos-2 model~\cite{peng2023kosmos}, an MLLM model that introduces novel capabilities for understanding object descriptions and linking text to visual elements. This enables the annotation of all images with MIP labels (bounding boxes) and corresponding explanations, highlighting the importance of each person.

\noindent \textbf{Label Refining Strategy.} Our label-refining strategy for annotating the Most Important Person (MIP) in images consists of two stages. In Stage 1, we use a MLLM to initially identify the MIP. The model receives the image and the prompt: ``\textless grounding\textgreater Most important person and its bbox value? \textless question\textgreater Why is this person important?" The MLLM then generates a bounding box (bbox) for the MIP and provides reasoning for their importance. In Stage 2, human annotators verify and classify the MLLM's annotations. They categorize the images into three groups: those where humans agree with the MLLM-identified MIP, those where both humans and the MLLM identify multiple MIPs (indicating the model's difficulty in selecting a single MIP), and those where humans disagree with the MLLM's choice. For images with disagreement or MLLM failure, manual annotation is performed using the VGG Annotator tool~\cite{vgg}. These human-annotated images are then re-evaluated by the MLLM with the prompt:``\textless question\textgreater Why the person in bbox is MIP?" This generates descriptions of the MIP. The MIP with majority agreement in each image is marked as the final MIP, and their descriptions are recorded as the final response.

\noindent \textbf{Analysis.} Figure~\ref{fig: teaser} illustrates a comparison of the outputs of the Kosmos-2~\cite{peng2023kosmos}. Here, the context information is highlighted as if it is a seminar audience (right image) or a winning celebration (left one). In the given context, the person holding the microphone or trophy is termed an MIP. 
% \vspace{-0.2 in}
\begin{table}[b]
\centering
\vspace{-0.2 in}
\caption{Number of subjects, subjective attributes in MIP-GAF.}
\scalebox{0.82}{
\begin{tabular}{l|ccccc}
\toprule[0.4mm]
\rowcolor{mygray} \textbf{Subset} & \textbf{\#Subjects} & \textbf{Mean Age} & \multicolumn{2}{c}{\textbf{Gender}} & \textbf{\#Images} \\
\rowcolor{mygray} \textbf{} & \textbf{} & \textbf{} & \textbf{Male} & \textbf{Female} & \textbf{} \\ \hline \hline
Train & 86,711 & 28.48 & 58,171 & 28,540  &  9,615 \\
Validation & 32,994& 28.51  & 21,453 & 11,541  &  4,113\\
Test & 26,840 &  28.28  & 18,057 & 8,783  & 2,822 \\ \hline
Overall & 146,545  & 28.42& 97,681 & 48,864   & 16,550  \\
\bottomrule[0.4mm]
\end{tabular}}
\label{tab:dataset_stats}
\vspace{-9mm}
\end{table}
% \vspace{-0.2 in}

% \subsection{Data Statistics}\label{sec:data_stat}

\noindent \textbf{Data Statistics}
We split the dataset into three non-overlapping sets: \textit{train}, \textit{validation}, and \textit{test} sets. In each set, we have the following statistics: 9,615 training images, 4,113 validation images, and 2,822 test images.Less than 8\% of the dataset images have MIP in the center of the frame. The overview of data distribution is presented in Table~\ref{tab:dataset_stats} and Figure~\ref{fig:data_stat_fig}. As the metadata of the MIP-GAF dataset does not contain gender and ethnicity distribution, we use off-the-shelf ~\href{https://github.com/vladmandic/face-api?tab=readme-ov-file}{Face-api} to infer age, gender, and ethnicity. 
% Among the detected faces, 68.9\% belongs to the young adult category. 

\begin{figure}[!ht]
    \centering
    \includegraphics[width=41mm,height=40mm]{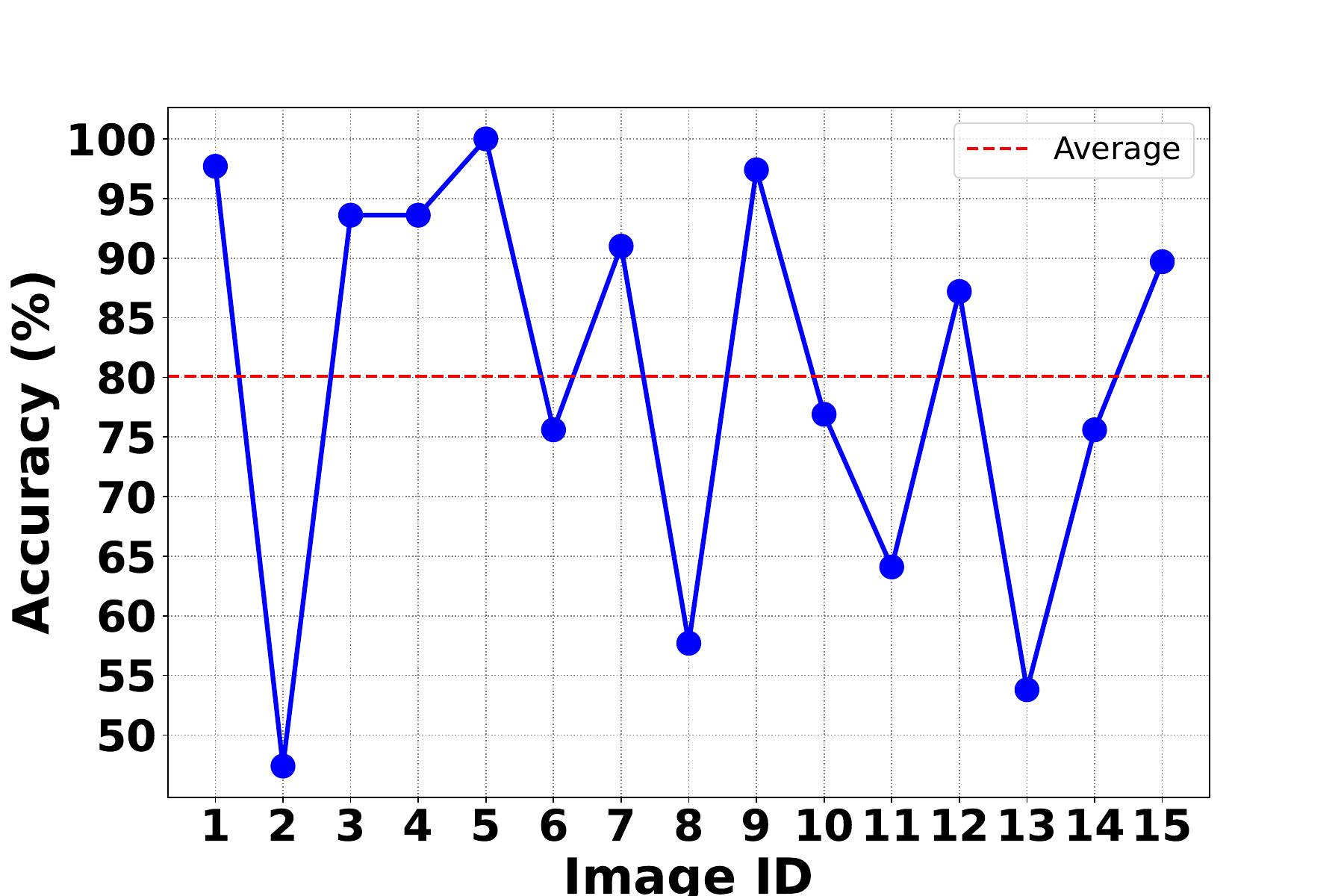}
    \includegraphics[width=41mm,height=36mm]{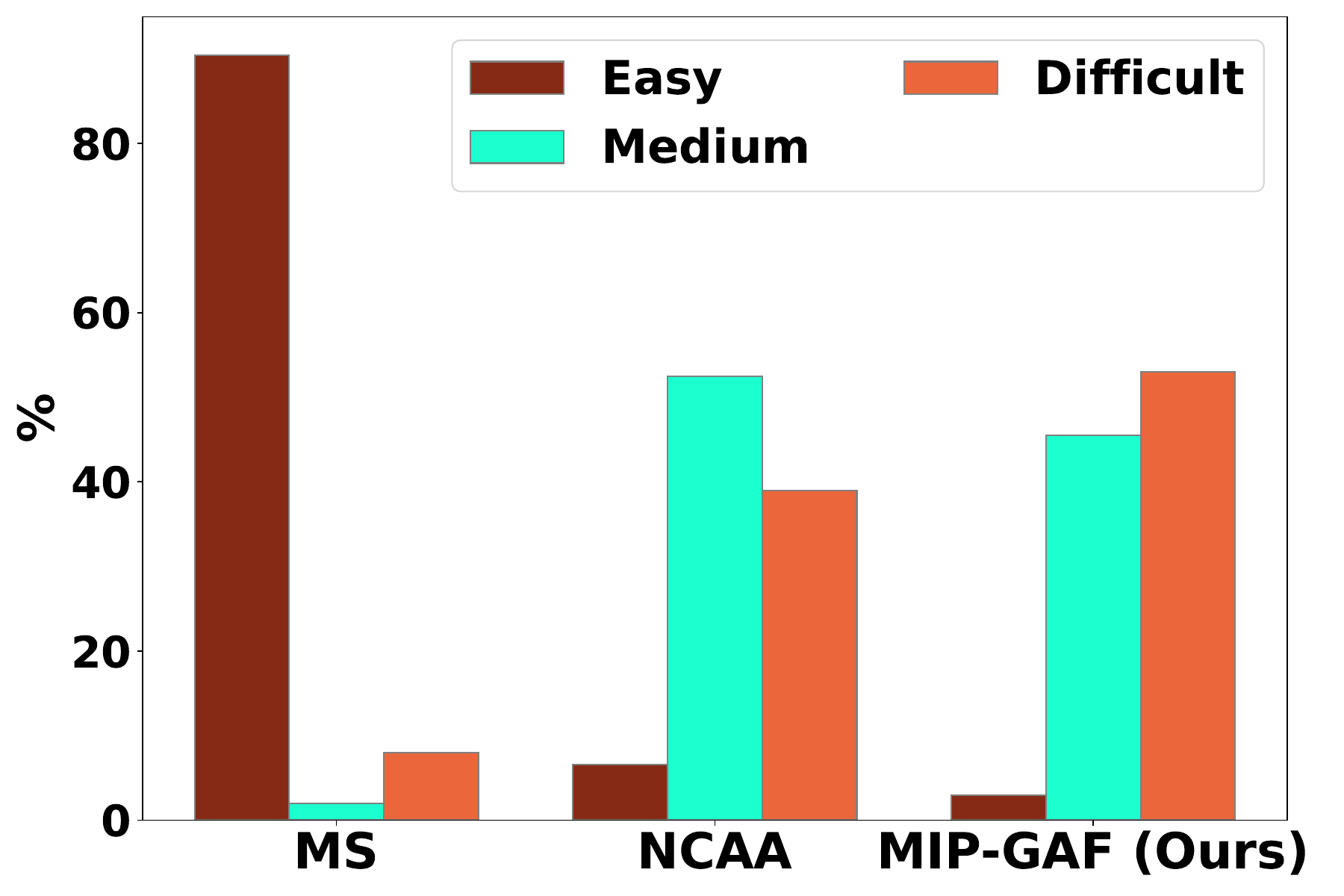}
    \caption{User study results. \textit{Left.} Human agreement analysis over images. \textit{Right.} We show the dataset-specific level of difficulty in spotting the MIP. The plot shows that the MIP is easily spottable for the MS dataset. Our proposed dataset, MIP-GAF, is more difficult than MS and NCAA.}
    \label{fig:user_study}
    % \vspace{-6mm}
    % \vspace{-9mm}
\end{figure}

\noindent \textbf{Data Quality Assessment}
To compare MLLM-based reasoning with human annotation, we conduct a user study with 78 participants (excluding the authors). Participants were shown 15 images and asked the question \textit{``Do you think the person in the bounding box is MIP in the given image?"}. The results, shown in Figure~\ref{fig:user_study}, indicate an 80\% agreement. Additionally, participants also rated the difficulty of identifying the MIP in three images from the MS, NCAA, and MIP-GAF datasets. This experiment is repeated for ten images, showing a difficulty trend of MS $<$ NCAA $<$ MIP-GAF. The results suggest our annotation pipeline meets human-level expectations.\\
\noindent \textbf{Agreement Analysis.} We have computed agreement among annotators (the MLLM and human raters) using Cohen's Kappa ($\kappa$) measure \cite{mchugh2012interrater}. To this end, we computed $\kappa$ between MLLM labels and human annotators, which is found to be 0.61. Specifically, out of an overall 16,650 instances, both MLLM and humans agreed on the common label in 10,600 cases, rejecting the incorrect label in 3,300 instances, while in 2,650 instances, human annotators did not agree with MLLM. This measure suggests that while individual differences exist in the perception of the most important person in the image, there is moderate to substantial agreement between the assessments of the MLLM and human annotators, implying that the considered images are effective for MIP detection.

\begin{figure*}[t]
    \centering
    \includegraphics[width=0.99\textwidth]{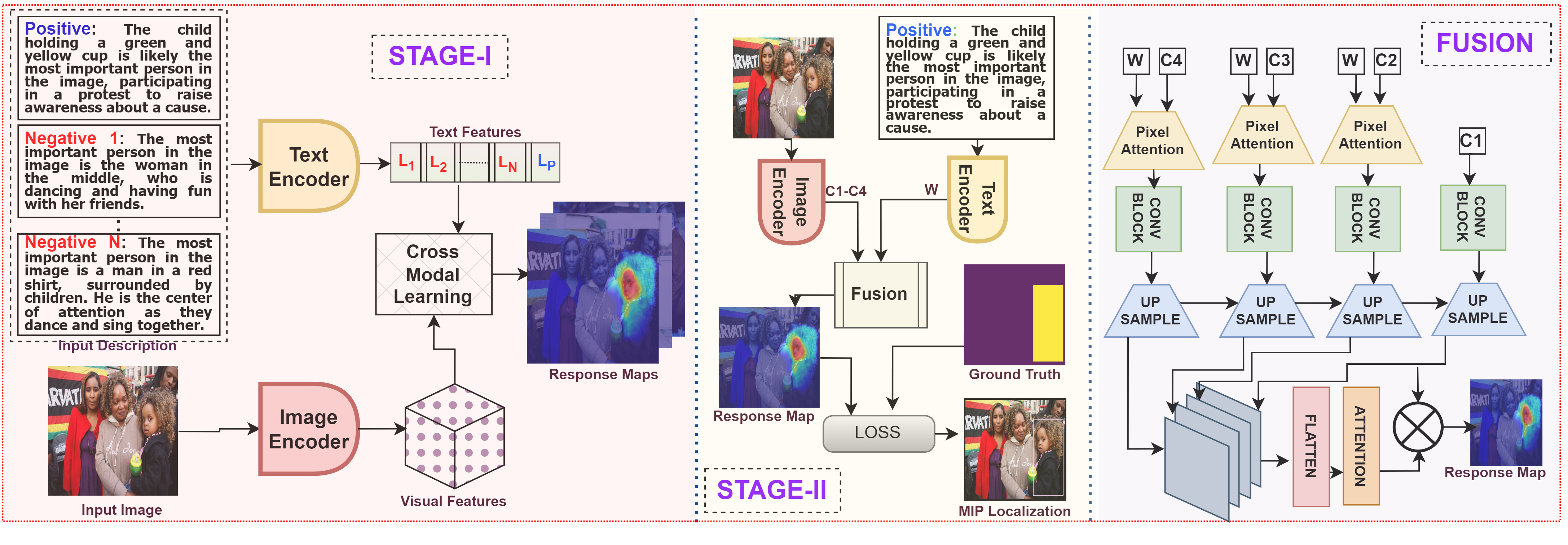}
    \vspace{-3mm}
    \caption{\textit{Our proposed MIP-CLIP framework.} Stage 1: It learns to classify text inputs and uses positive expressions to locate the MIP on response maps. Stage 2: Trained image and text encoders generate feature maps, and a \textbf{fusion} model localizes MIP using response maps.}
    \label{fig:baseline}
    \vspace{-7mm}
\end{figure*}

% \subsection{}
 
% This level of agreement supports the validity of the annotated data for subsequent analysis. 
% \sg{we need a figure for this}

\section{Proposed Method: MIP-CLIP}
Overall framework for MIP-CLIP is shown in Figure~\ref{fig:baseline}. The primary goal of proposed MIP localization is to establish pixel-level correspondence between visual content and referring descriptions without pixel-level annotations, using a limited supervised approach. Following~\cite{image_segmentation}, our model operates in two stages. In stage 1, the model classifies positive and negative expressions for each input image while localizing the MIP described by the positive description. Positive description represents the MIP in the input image, while negative descriptions are from other images. This classification process models text-to-image responses, associating the input image's visual content with the positive expressions.
Stage 2 uses the trained image and text encoders from stage 1 to generate corresponding embedding, which are passed to a \textbf{fusion block} to create the final response map. This response map is compared with the ground truth map, and the model learns to localize the MIP by optimizing pixel-wise binary cross-entropy loss.
The fusion block processes text (W) and image (C1-C4) embeddings through a pixel-wise attention layer to identify relevant pixels. The attention embedding goes through a convolutional block (with Conv2D, batch normalization, and PReLU activation) and an upsampling layer. This upsampling layer integrates information from C4 to C1, ensuring C1 contains all crucial information. The refined upsampled feature map (C1) highlights specific MIP information. All upsampled maps (C1-C4) are stacked, flattened, and passed to an attention layer (a linear layer with softmax activation)  to learn specific details from each map, generating attention weights for each pixel. These weights are multiplied with upsampled C1 to determine pixel significance, forming the final response map to compare with the ground truth. Modified ResNet~\cite{resnet} is used as the image encoder, and CLIP's BERT~\cite{clip_bert} as the text encoder. During prediction, descriptions containing multiple sentences are split at full stops, generating an output for each sentence. The ReLU function is applied to each sentence's final output, and their sum becomes the final predicted response map. 
% This response map is used to identify disjoint areas, from which bounding boxes are calculated. The bounding box with the largest area is considered the MIP.

\section{Experiments}
\begin{figure*}[!h]
    \includegraphics[scale=0.65]{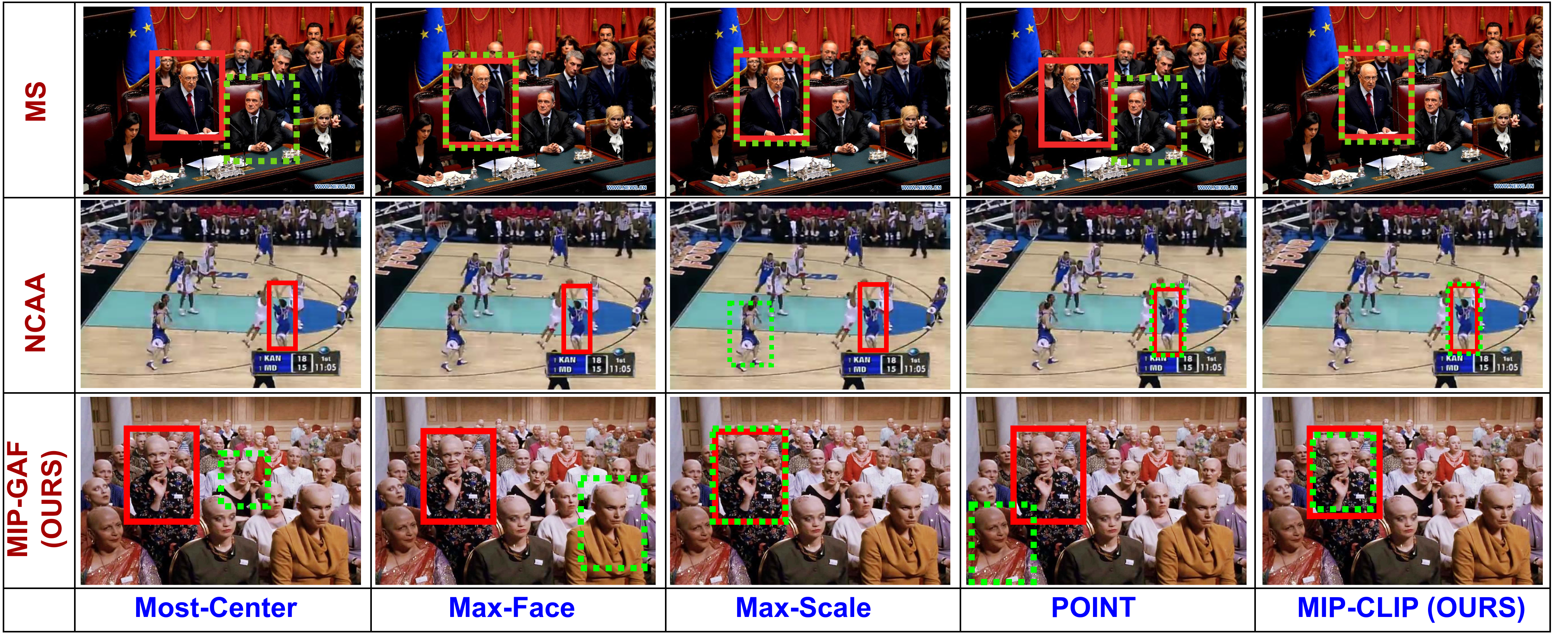}
    \caption{\textit{Qualitative Analysis.} We compare the output of different off-the-shelf methods on MS, NCAA, and MIP-GAF datasets. Here, the dotted line(green) indicates the predicted bounding box and the solid line (red) bounding box indicates the ground truth.}
    \label{fig:qualitative}
    \vspace{-8mm}
\end{figure*}

\noindent \textbf{Existing Benchmarks.} The \textbf{MS} Dataset includes 2,310 images (training: 924, validation: 232, testing: 1,154) with person-specific importance labels and face bounding boxes across six scene types. The \textbf{NCAA} Dataset has 9,736 basketball images with person-level bounding boxes and importance annotations. The \textbf{Extended MS (EMS)} Dataset contains 10,687 images (training: 8,607 with 690 labeled, validation: 230 labeled, testing: 1,390 labeled). The \textbf{Extended NCAA (ENCAA)} Dataset includes 19,062 sports images (training: 2,825 labeled, validation: 941 labeled, testing: 5,970 labeled).

\noindent \textbf{Experimental Protocols.}
We compared our model with several baseline models outlined in the literature below:

\noindent \textit{1. Most-Center:} the person closest to the image center.

\noindent \textit{2. Max-Scale:} the person with the largest area in the image

% the person having a larger area (face/body) in an image.

\noindent \textit{3. Max-Face:} person with the largest visible face.

\noindent \textit{4. Max-Saliency:} We investigate the correlation between the salient regions in an image concerning the MIP.

\noindent \textit{5. POINT:} POINT~\cite{li2019learning} is a deep relation-based framework that learns to build the interpersonal relationship modeling with feature learning for MIP localization.

\noindent \textit{6. Semi-POINT:} Semi-POINT~\cite{hong2020learning} framework aims to assign pseudo-labels to individuals in un-annotated images. Upon assigning pseudo labels, the model learns to update the MIP localization model based on both labels and pseudo-labels.

\noindent \textit{7. CLIP:} We fine-tuned the CLIP~\cite{radford2021learning} model using MIP-GAF images and their corresponding descriptions. Additionally, we utilize the vision encoder to extract features and incorporate an MLP module consisting of two dense layers with sizes 512 and 256, followed by an output layer with four neurons to localize the MIP coordinates.

% We fine tune the CLIP~\cite{radford2021learning} model with MIP-GAF images and corresponding descriptions. Further, we use the vision encoder as feature extractor and added a MLP module containing two dense layer of size 512 and 256 followed by output layer containing 4 neuros to predict the MIP coordinates.

\noindent \textit{8. Zero-shot MLLM:} We use off-the-shelf multimodal large language models such as CogVLM~\cite{hong2023cogagent,wang2023cogvlm} and QwenVL~\cite{Qwen-VL} to see the performance in zero shot inference.

\noindent \textbf{Level of Supervision.} For benchmarking and evaluation, we use four levels of supervision: Zero-shot learning adapts pre-trained weights of large language models (CogVLM and QwenVL) directly to the MIP-GAF test set. Fully supervised learning involves training models on the train set and evaluating them on validation and test sets, applied to most-center, max-scale, max-face, max-saliency, and POINT protocols. Semi-supervised learning uses 33\% and 66\% labeled data from the training partition, following~\cite{li2019learning}, for training and evaluates on validation and test sets. Self-supervised learning employs vision-language self-supervision, pre-training with MIP-GAF images and descriptions, followed by downstream adaptation.

% \noindent \textbf{Level of supervision.} For benchmarking and evaluation purposes, we use four different levels of supervision described below: 

% \noindent \textit{1. Zero-shot learning.} We use the pre-trained weights of the off-the-shelf large language model (CogVLM~\cite{hong2023cogagent,wang2023cogvlm} and QwenVL~\cite{Qwen-VL}) to adapt to the MIP-GAF's test set directly.

% \noindent \textit{2. Fully Supervised.} In this protocol, we use train set data to train the respective model, val, and test set data for evaluation purposes. We use a fully supervised strategy for most-center, max-scale, max-face, max-saliency, and POINT protocols.

% \noindent \textit{3. Semi-Supervised.} In this protocol, we utilize 33\% and 66\% labels from training partition following~\cite{li2019learning} for model training. Further, evaluation is performed in validation and test sets.

% \noindent \textit{4. Self Supervised.} In this protocol, we use the Vision language self-supervision technique, i.e., pre-trained with the MIP-GAF images along with the corresponding description. After pre-training stage, we perform downstream adaptation to MIP using linear probing.

% \subsection{Evaluation Metrics}

\noindent \textbf{Evaluation Metrics.}
% We follow the standard evaluation protocol following the prior works~\cite{li2019learning}. To quantify the performance of the methods above on MIP Localization, the mean Average Precision (mAP) is used to assess the correctness of detected MIP. Following the POINT~\cite{li2019learning} framework, the cumulative matching characteristics (CMC) curve is also reported in our paper to show the results of the top k-rank important person.
We follow the standard evaluation protocol from prior works~\cite{li2019learning}. Mean Average Precision (mAP) is used to measure MIP Localization performance. Following the POINT framework~\cite{li2019learning}, we also report the cumulative matching characteristics (CMC) curve to show the top k-rank important persons.

\begin{figure}[t]
    \centering
    \includegraphics[width=85mm,height=30mm]{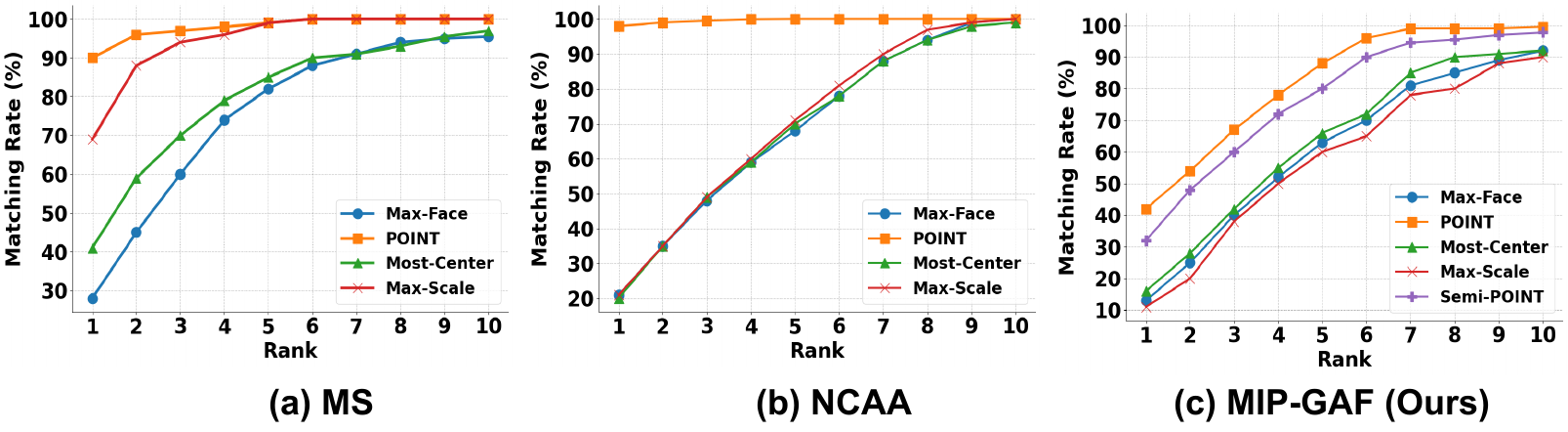}
    % \vspace{-3mm}
    % \caption{Cumulative Matching Characteristics (CMC) Curve on (a) MS, (b) NCAA, and (c) MIP-GAF datasets. The curve directly shows that for MS and NCAA, due to simplistic scenes, the POINT~\cite{li2019learning} framework can identify MIP with a matching rate $\geq 90\%$. Whereas in our proposed dataset, it is $\sim 40\%$. Please see Section~\ref{sec:qualitative} for more details.}
    \caption{CMC curves for (a) MS, (b) NCAA, and (c) MIP-GAF datasets. The POINT framework~\cite{li2019learning} achieves $\geq 90\%$ matching rates for MS and NCAA, reflecting their simplistic scenes. In contrast, on our MIP-GAF dataset, the rate is $\sim 40\%$. See Section~\ref{sec:qualitative} for details.}\label{fig:cmc}
    \vspace{-7mm}
\end{figure}

\noindent \textbf{Implementation details.}
We implemented the experimental protocols on PyTorch~\cite{paszkePyTorch2019} with Nvidia A100 40GB GPU. We tried our best to incorporate the off-the-shelf methods from the GitHub repositories. 
\noindent We trained both the POINT~\cite{li2019learning} and semi-POINT~\cite{hong2020learning} model for 200 epochs with early stopping having patience value as 5. We adopt the same settings as the public GitHub repository for hyperparameters. We observed that both the models have relatively the same configuration setup where the hyper-parameters are tuned on the validation set of the data~\cite{li2019learning,hong2020learning}. 
% The weight decay is 0.0005, and the momentum is 0.5 in all experiments. The learning rate is initialized to 0.002, and we follow the learning rate update strategy mentioned in literature~\cite{li2018personrank} i.e., the learning rate is scaled by a factor of 0.5 every 20 epochs. For both cases, SGD is used as an optimizer~\cite{zinkevich2010parallelized}.
\noindent During the pre-training process of the CLIP model~\cite{radford2021learning}, our model's weight was initialized with CLIP's default weight. We pre-train the CLIP model on the training partition of MIP-GAF by Adam optimizer~\cite{kingma2014adam} with a learning rate of $5e-5$, where $\beta$ value ranges in ($0.9$, $0.98$). To prevent the condition of division by zero, the $\epsilon$ value is set to $1e-6$ along with a weight decay of $0.2$. The model is trained for 20 epochs for downstream linear probing. To incorporate the most important person localization aspect, we have employed the $\ell_1$ loss function in the prediction stage with the ReLU activation function. To train the proposed MIP-CLIP method, stage 1 uses classification and contrastive losses, while stage 2 employs binary cross-entropy loss with the Adam optimizer at a learning rate of 0.0005.  The model is trained for 30 epochs with early stopping (patience of 3), and the best model is used for further evaluation. In all the above cases, we use the standard evaluation metric from~\cite{li2018personrank,li2019learning,hong2020learning}, reporting mean average precision to measure performance. We also include a CMC graph to show the ranking process (see Figure~\ref{fig:cmc}).

\section{Results and Analysis}
\subsection{Quantitative Analysis}
This section comprises the quantitative results obtained
using the state-of-the-art MIP localization methods when trained and evaluated on the MIP-GAF dataset.

\begin{table}[t]
\caption{\textbf{MIP Detection Benchmarks.} We compare the state-of-the-art models on our proposed MIP-GAF dataset.}
\label{tab:sota_results}
\scalebox{0.85}{
\begin{tabular}{l|l||l|c|c}

\toprule[0.4mm]
\rowcolor{mygray}\textbf{Data} &\textbf{Supervision} &  \textbf{Method} & \multicolumn{1}{c|}{\textbf{Val}} & \multicolumn{1}{c}{\textbf{Test}} \\
\rowcolor{mygray}\textbf{} & \textbf{} & \textbf{} & \textbf{mAP $\uparrow$} & \textbf{mAP$\uparrow$} \\
\hline\hline
 \multirow{10}{*}{\rotatebox[origin=c]{90}{\textbf{MIP-GAF}}} &  \multirow{2}{*}{Zero-shot}& Qwen-VL~\cite{Qwen-VL} & 20.76 & 14.56 \\ \cline{3-5}
&   & CogVLM~\cite{wang2023cogvlm, hong2023cogagent} & 13.44 & 13.87 \\ \cline{2-5}
 &\multirow{5}{*}{Supervised} & Most-Center & 49.17 &23.19 \\ \cline{3-5}
 &   &  Max-Scale & 13.31  & 12.50\\ \cline{3-5}
 &   & Max-Face &  42.52& 17.79\\ \cline{3-5}
 &   &  Max-Saliency & 11.50 & 12.25\\ \cline{3-5}
  &   &  POINT~\cite{li2019learning} & 88.70 & 72.79\\ \cline{2-5}
 &   \multirow{4}{*}{Semi-Supervised} & semi-POINT~\cite{hong2020learning} & \multirow{2}{*}{67.13} & \multirow{2}{*}{63.85} \\ 
 & & (Labelled: 33\%) & & \\ \cline{3-5}
  &   & semi-POINT~\cite{hong2020learning} & \multirow{2}{*}{69.93} & \multirow{2}{*}{65.05}\\ 
   & & (Labelled: 66\%) & & \\ \cline{2-5}
   
 &   \multirow{2}{*}{Self-Supervised} & CLIP~\cite{radford2021learning} & 43.83 & 45.67\\ \cline{3-5}
&& MIP-CLIP & \textbf{74.73}   & \textbf{71.92}\\ 
\bottomrule[0.4mm]
\end{tabular}}
\vspace{-6mm}
\end{table}

\noindent \textbf{Comparision with State-of-the-Art Benchmarks.} The results in Table~\ref{tab:sota_results} reveal that models developed on previous datasets perform poorly on the MIP-GAF dataset. For zero-shot benchmarking, using MLLMs like Cog-VLM~\cite{wang2023cogvlm,hong2023cogagent} and Qwen-VL~\cite{Qwen-VL} shows existing algorithms struggle with scene-based inference. In supervised benchmarks, the test mAP scores for most center (23.19), max scale (12.50), max-face (17.79), and max-saliency (12.25) are notably low, highlighting the richness of social context and diversity in the MIP-GAF. The semi-supervised benchmark further confirms this pattern, with test mAP dropping to 63.85 and 65.05 at 33\% and 66\% labeled data, respectively. In self-supervised learning with the CLIP model~\cite{radford2021learning}, the validation and test set mAP drop significantly to 43.83 and 45.67, while the proposed MIP-CLIP achieves results comparable to the supervised approach (POINT), highlighting the dataset's suitability for 'in-the-wild' conditions.
\begin{table}[!h]
\vspace{-0.2 in}
\caption{\textbf{Benchmark Comparison.} We compare the state-of-the-art methods on MS, NCAA, EMS and ENCAA datasets.}
\label{tab:crossdata_results}
\scalebox{0.8}{
\begin{tabular}{l||l|l|c|c}
\toprule[0.4mm]
\rowcolor{mygray}\textbf{Supervision} & \textbf{Dataset} & \textbf{Method} & \multicolumn{1}{c|}{\textbf{Val}} & \multicolumn{1}{c}{\textbf{Test}} \\
\rowcolor{mygray}\textbf{} & \textbf{} & \textbf{} & \textbf{mAP $\uparrow$} & \textbf{mAP$\uparrow$} \\
\hline\hline
\multirow{4}{*}{Zero-shot} & MS & \multirow{3}{*}{Qwen-VL~\cite{Qwen-VL}} & 16.24 & 12.72\\ \cline{4-5}
& NCAA &  & 39.81  & 25.95 \\ \cline{4-5}
% & CUC &  &  & \\ \cline{4-5}
 & MIP-GAF &  &20.76  &14.56 \\ \cline{2-5}
& MS & \multirow{3}{*}{Cog-VLM~\cite{wang2023cogvlm,hong2023cogagent}} & 15.87 & 16.21 \\ \cline{4-5}
& NCAA & &30.20 & 32.39   \\ \cline{4-5}
% & CUC &  &  & \\ \cline{4-5}
 & MIP-GAF & &13.44 & 13.87  \\ \hline
 % \multirow{5}{*}{Supervised} & MS & \multirow{4}{*}{POINT~\cite{li2019learning}} & - &92.00 \\ \cline{4-5}
 % & NCAA & & - &97.30 \\ \cline{4-5}
 % % & CUC &  & - &49.50 \\ \cline{4-5}
 % & MIP-GAF &  & 88.70 &72.79 \\ \hline 
 \multirow{5}{*}{Supervised} & MS & \multirow{4}{*}{POINT~\cite{li2019learning}} & - &92.00 \\ \cline{4-5}
 & NCAA & & - &\textbf{97.30} \\ \cline{4-5}
 % & CUC &  & - &49.50 \\ \cline{4-5}
 & MIP-GAF &  & 88.70 &72.79 \\ \hline
\multirow{4}{*}{Semi-Supervised} & EMS & \multirow{3}{*}{semi-POINT~\cite{hong2020learning}} & - & 87.81 \\ \cline{4-5}
& ENCAA &  & - &88.75 \\ \cline{4-5}
% & CUC &  &  & \\ \cline{4-5}
 & MIP-GAF & (Labelled: 33\%)  & 67.13 &63.85 \\ \cline{2-5}
& EMS & \multirow{3}{*}{semi-POINT~\cite{hong2020learning}} & - &88.44 \\ \cline{4-5}
& ENCAA &  & - &90.86 \\ \cline{4-5}
% & CUC &  &  & \\ \cline{4-5}
 & MIP-GAF & (Labelled: 66\%) & 69.93 & 65.05\\ \hline
\multirow{4}{*}{Self-Supervised} & MS & \multirow{3}{*}{CLIP~\cite{radford2021learning}} & 24.56 & 22.18 \\ \cline{4-5}
& NCAA &  & 29.64 &31.88 \\ \cline{4-5}
% & CUC &  &  & \\ \cline{4-5}
 & MIP-GAF &   & 43.83 &45.67 \\ \cline{2-5}
& MS & \multirow{3}{*}{MIP-CLIP} & 81.40 &84.00 \\ \cline{4-5}
& NCAA &  & 15.80 &16.20 \\ \cline{4-5}
% & CUC &  &  & \\ \cline{4-5}
 & MIP-GAF &  & 74.73 & 71.92\\ 
 % \hline 
 % \multirow{1}{*}{Self-Supervised} & MS & \multirow{3}{*}{CLIP~\cite{radford2021learning}} & 24.56 & 22.18\\ \cline{4-5}
 % & NCAA &  &29.64  & 31.88\\ \cline{4-5}
 % % & CUC & CLIP &  & & & & &\\ \cline{3-9}
 % & MIP-GAF &  & 43.83 & 45.67\\ 
 % % \multirow{1}{*}{Causality} &  & Scene-Graph &  & & & & &\\ 
\bottomrule[0.4mm]
\end{tabular}}
\end{table}
% \vspace{-0.15 in}

\noindent \textbf{Dataset Benchmark Comparison.}
We have also conducted experiments to compare the performance with the existing benchmark datasets:MS~\cite{li2018personrank}, Extended MS (EMS)~\cite{li2019learning}, NCAA~\cite{li2018personrank} and Extended NCAA~\cite{li2019learning}. The results are shown in Table~\ref{tab:crossdata_results}. For the supervised learning framework POINT~\cite{li2019learning}, the mAP performance drops to 72.79 as compared with MS (92.00) and NCAA (97.30). The performance seems saturated for MS and NCAA datasets as the MIPs are in either frontal view or fall under the salient region. Whereas for other learning paradigms, it's still an open avenue to explore.

\noindent \textbf{Transfer Learning over Datasets.} Additionally, we conducted a cross-dataset transfer learning experiment~\cite{weiss2016survey} using MIP-CLIP to assess the richness of the latent feature space. We train the model on one benchmark dataset and tested it on another, with results shown in Table~\ref{tab:transfer}. Our MIP-CLIP method, trained on the MIP-GAF dataset, outperforms those trained on other datasets including CLIP~\cite{radford2021learning}, Zero-shot~\cite{wang2023cogvlm,hong2023cogagent, Qwen-VL}, and achieves comparable results to the supervised method POINT~\cite{li2019learning} (see Table \ref{tab:sota_results}).
\begin{table}[t]
\centering

\caption{\textbf{Cross-Dataset Transfer Learning Results. MIP-CLIP trained on MIP-GAF outperforms other datasets, highlighting the richness of its latent feature space.}} 
\vspace{- 0.1in}
%\textmd{\underline{\small{\textit{cross data}}. 
\label{tab:transfer}
\scalebox{0.8}{
\begin{tabular}{cc||cc}
\toprule[0.4mm]
\rowcolor{mygray} \multicolumn{2}{r||}{\textbf{Methods $\rightarrow$}} & \textbf{MIP-CLIP} & \textbf{}\\
\rowcolor{mygray} \textbf{Train Data} & \textbf{Test Data} &\textbf{Val mAP $\uparrow$} & \textbf{Test mAP $\uparrow$} \\ \hline \hline
MS & MIP-GAF & 76.20  & 72.5   \\
NCAA & MIP-GAF & 40.00 & 40.50  \\ 
\hline
% CUC & MIP-GAF &  &  \\
MIP-GAF & MS & \textbf{89.20}& \textbf{88.80} \\
MIP-GAF & NCAA & 9.00 & 9.00  \\ 
% \hline
% MS & NCAA & 22.55 & 26.15  \\
% NCAA & MS & 28.11 & 29.50  \\
% MIP-GAF & CUC &  &  \\
\bottomrule[0.4mm]
\vspace{-10mm}
\end{tabular}}
\end{table}

% Additionally, we have performed a cross-dataset experiment on transfer learning~\cite{weiss2016survey} using MIP-CLIP to get an overview of richness in terms of latent feature space. We train the model on one benchmark dataset and test it on another benchmark dataset. The results are shown in Table~\ref{tab:transfer}. It shows that our MIP-CLIP method trained on our MIP-GAF dataset performs better than those trained on other datasets.

% \noindent \textbf{Group Emotion Recognition Comparison.} We have also performed group emotion recognition as a downstream task inspired by~\cite{xie2023most}. There is a significant increase in performance which indicates that the context and explainability aspect helps to decode different aspects of social group interactions.   

\subsection{Qualitative Analysis}
\label{sec:qualitative}
% \subsubsection{Visualization of Output Bounding Box} 

\noindent \textbf{Visualization of Output Bounding Box.} The performance comparison of different state-of-the-art models on MIP-GAF, MS, and NCAA dataset are shown in Figure~\ref{fig:qualitative}.
% Here, the dotted line indicates the predicted bounding box and the solid line bounding box indicates the ground truth. 
The results indicate that our dataset is more challenging and requires a more robust algorithm.

% \subsubsection{CMC Graph} 
\noindent \textbf{CMC Graph.} We plot the Cumulative Matching Characteristics (CMC) curves following~\cite{li2018personrank,li2019learning} of different methods on MS, NCAA, and MIP-GAF datasets (See Figure~\ref{fig:cmc}). These graphs compare state-of-the-art methods. The figures show that for the MS and NCAA datasets, the POINT framework~\cite{li2019learning} can identify MIPs with a matching rate of $\geq 90\%$, attributed to the simplistic scene information. It indicates that existing SOTA needs to be more robust for `in-the-wild' scenes. We believe that our dataset will be a valuable asset to the research community.

% These graphs compare state-of-the-art methods, the results reported in the figures show that for MS and NCAA, due to simplistic scene information, POINT~\cite{li2019learning} framework able to identify MIP with a matching rate $\geq 90\%$. Whereas in our proposed dataset MIP-GAP the performance drops to $\sim 40\%$. It directly indicates that existing SOTA need to be more robust for `in-the-wild' scenes. We believe that our dataset will be a valuable asset to the research community.

%-------------------------------------------------------------------------
\section{Conclusion}
This paper presents MIP-GAF, a large-scale dataset for the most important person localization. Our proposed semi-automatic data labeling paradigm utilizes the power of MLLM to annotate the context-level situation understanding aspect. The comprehensive benchmarking of the dataset using state-of-the-art methods indicates a significant drop in performance. This indicates that the proposed dataset will play a crucial role in the MIP research area for algorithm development. 
\noindent \textbf{Broader Impact.} We believe that MIP-GAF can be an important benchmark for the multimedia community for aiding researchers in developing algorithms on human-human interaction `in the wild'. Owing to the rich, explainable information, it would be easy to get more context information for real-world applications.
\noindent \textbf{Limitations.} Potential bias can be introduced in the model as we use the existing face detection library~\cite{Vladmandic}. We will eliminate these limitations in our updated versions. 

%-------------------------------------------------------------------------

%------------------------------------------------------------------------
%%%%%%%%% REFERENCES
{\small
\bibliographystyle{ieee_fullname}
\bibliography{egbib}
}

\end{document}